\documentclass[sigconf]{acmart}

\AtBeginDocument{%
  \providecommand\BibTeX{{%
    \normalfont B\kern-0.5em{\scshape i\kern-0.25em b}\kern-0.8em\TeX}}}

\copyrightyear{2023} 
\acmYear{2023} 
\setcopyright{acmlicensed}\acmConference[GeoSearch '23]{2nd ACM SIGSPATIAL International Workshop on Searching and Mining Large Collections of Geospatial Data}{November 13, 2023}{Hamburg, Germany}
\acmBooktitle{2nd ACM SIGSPATIAL International Workshop on Searching and Mining Large Collections of Geospatial Data (GeoSearch '23), November 13, 2023, Hamburg, Germany}
\acmPrice{15.00}
\acmDOI{10.1145/3615890.3628527}
\acmISBN{979-8-4007-0352-2/23/11}

\usepackage{xcolor}
\usepackage{wrapfig}
\usepackage{multirow}
\usepackage{hyperref}
\usepackage{soul}

\begin{document}

\title{A Critical Perceptual Pre-trained Model for
Complex Trajectory Recovery}

\author{Dedong Li}
\email{ddlecnu@gmail.com}
\affiliation{%
  \institution{SenseTime Research}
  \city{Shanghai}
  \country{China}
}

\author{Ziyue Li}
\authornote{Corresponding Author.}
\email{zlibn@wiso.uni-koeln.de}
\orcid{0000-0003-4983-9352}
\affiliation{%
  \institution{University of Cologne}
  \streetaddress{Albertus-Magnus-Platz}
  \city{Cologne}
  \state{NRW}
  \country{Germany}
  \postcode{50923}
}

\author{Zhishuai Li}
\email{lizhishuai@sensetime.com}
\affiliation{%
  \institution{SenseTime Research}
  \city{Shanghai}
  \country{China}}

\author{Lei Bai}
\email{baisanshi@gmail.com}
\affiliation{%
 \institution{Shanghai AI Laboratory}
 \streetaddress{No. 197 Yueyang Rd}
 \city{Shanghai}
 \country{China}}

\author{Qingyuan Gong}
\email{gongqingyuan@fudan.edu.cn}
\affiliation{%
 \institution{Fudan University}
 \streetaddress{No. 220 Handan Rd}
 \city{Shanghai}
 \country{China}}

\author{Lijun Sun}
\email{lijun.sun@mcgill.ca}
\orcid{0000-0001-9488-0712}
\affiliation{%
  \institution{McGill University}
  \streetaddress{845 Sherbrooke St W}
  \city{Montreal}
  \state{Quebec H3A 0G4}
  \country{Canada}
}

\author{Wolfgang Ketter}
\email{ketter@wiso.uni-koeln.de}
\orcid{0000-0001-9008-142X}
\affiliation{%
  \institution{University of Cologne}
  \streetaddress{Albertus-Magnus-Platz}
  \city{Cologne}
  \state{NRW}
  \country{Germany}
  \postcode{50923}
}

\author{Rui Zhao}
\email{zhaorui@sensetime.com}
\orcid{0000-0001-5874-131X}
\affiliation{%
  \institution{SenseTime Research}
  \city{Shanghai}
  \country{China}}
\renewcommand{\shortauthors}{Li and Li, et al.}

\begin{abstract}
Trajectory on the road traffic is commonly collected at a low sampling rate, and trajectory recovery aims to recover a complete and continuous trajectory from the sparse and discrete inputs. Recently, sequential language models have been innovatively adopted for trajectory recovery in a pre-trained manner: it learns road segment representation vectors, which will be used in the downstream tasks. However, existing methods are incapable of handling \textit{complex} trajectories: when the trajectory crosses remote road segments or makes several turns, which we call \textit{critical nodes}, the quality of learned representations deteriorates, and the recovered trajectories skip the critical nodes. This work is dedicated to offering a more robust trajectory recovery for complex trajectories. Firstly, we define the trajectory complexity based on the detour score and entropy score, and construct the complexity-aware semantic graphs correspondingly. Then, we propose a \textbf{M}ulti-view \textbf{G}raph and \textbf{C}omplexity \textbf{A}ware \textbf{T}ransformer (MGCAT) model to encode these semantics in trajectory pre-training from two aspects: 1) adaptively aggregate the multi-view graph features considering trajectory pattern, and 2) higher attention to critical nodes in a complex trajectory. Such that, our MGCAT is perceptual when handling the critical scenario of complex trajectories. Extensive experiments are conducted on large-scale datasets. The results prove that our method learns better representations for trajectory recovery, with \textbf{5.22\%} higher F1-score overall and \textbf{8.16\%} higher F1-score for complex trajectories particularly. The code is available  \href{https://github.com/bonaldli/ComplexTraj}{\textcolor{teal}{here}}.
\end{abstract}




\maketitle

\section{Introduction}
\begin{figure}[t]
\centering
\includegraphics[width=0.99\columnwidth]{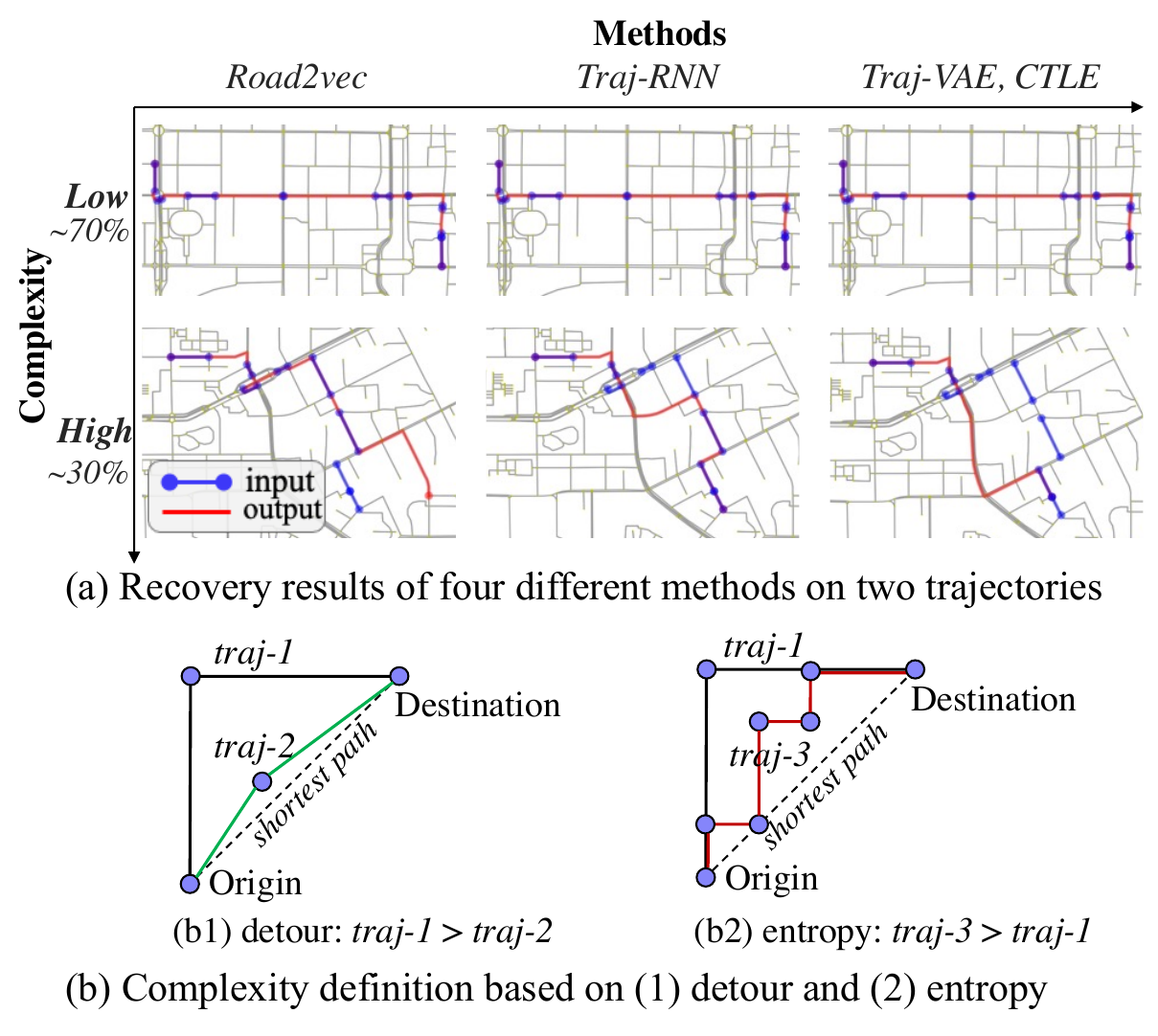}
\caption{(a) Recovery results of four different trajectory models (Road2Vec, Traj-RNN, Traj-VAE, and CTLE), with blue segments as the sparse and discrete input, and red line as the recovered trajectory output; (b) Trajectory complexity definition with detour and entropy. Detour: \textit{traj-3} $=$ \textit{traj-1} $>$ \textit{traj-2}. Entropy (zig-zag): \textit{traj-3} $>$ \textit{traj-1} $=$ \textit{traj-2}.}
\label{fig: motivation}
\end{figure}

Global  Positioning System (GPS) has been widely deployed in the navigation service of various devices. However, GPS records are commonly collected at a low sampling rate to reduce energy consumption \cite{MTrajRec}. This renders the GPS records of a trip sparse and discrete, shown as the blue inputs in Fig. \ref{fig: motivation}. (a). Before mining any valuable information from the low-sampling-rate trajectories based on GPS records, trajectory recovery is a necessary step. However, due to the high hardware cost, the cameras are sparsely deployed on road networks, rendering the input of locations also sparse.

Trajectory recovery is a fundamental task in Intelligent Transport Systems (ITS), which aims to recover a complete and continuous trajectory from sparse and discrete input road segments from GPS records \cite{wu:17, MTrajRec, xu2017trajectory, mao2022jointly}. Trajectory recovery further enables various applications such as urban movement behavior study \cite{9547354, li2021urban,ziyue2021tensor,9737052, li2022individualized}, traffic prediction \cite{w:19, f:20, li2020long, li2020tensor}, next location prediction \cite{l:21, zhang2022trajectory, li2022individualized}, route planning \cite{wu2020learning}, anomaly detection \cite{li2022profile}, and travel time estimation \cite{f:20, yang2021unsupervised}. 

Earlier, researchers proposed heuristic search \cite{zheng2015trajectory} and probabilistic models such as hidden Markov model \cite{newson2009hidden} to conduct trajectory recovery, {where the noisy location measurements are matched to the most likely states, \textit{i.e.}, road segments, such that a route is learned.} Recently, trajectory modeling  \cite{wu:17,zhou2018trajectory,zhou2020semi,fl:20,l:21} has especially received much attention. Language sequential models such as RNN, BERT, and VAE \cite{d:18, lewis2020bart, dong2019unified, liu2019roberta,zhou2018trajectory} have been adopted into the trajectory domain, treating a road segment as a token and a trajectory as a sequence, so as to learn a representation for the token from trajectories in a pre-training paradigm, which achieves well-recognized high generality over the statistical methods.

The models mentioned above have proved that considering the trajectory \textit{contexts} helps to learn a better road segment representation \cite{wu:17,zhou2018trajectory,fl:20,l:21}, where the context of a given road segment $s_{i}$ is deemed as the road segments that come before ($s_{<i}$) and after ($s_{>i}$) along trajectories. {Intuitively, the trajectory context encourages the road segments close along the trajectory to have similar representations}. Among them, CTLE \cite{l:21} further proposed to consider temporal dynamics, assuming that the contexts of road segments change over time since the same locations could serve different functions at different times.

However, merely considering the contexts in the trajectory might be inadequate, as we observed in experiments that the performances of the existing trajectory models deteriorate sharply when the trajectory becomes more complex \cite{w:19}. Fig. \ref{fig: motivation} demonstrates the trajectory recovery results from four state-of-the-art models \cite{wu:17,zhou2018trajectory,fl:20,l:21}. 
When handling low-complexity trajectories, as shown in the upper row of Fig. \ref{fig: motivation}. (a), the four methods have equally good recovery results. However, when dealing with complex trajectories, these models tend to make a common mistake, \textit{i.e.}, skipping some input road segments that are far or involve turns (for short, we call them ``critical nodes''), as shown in the bottom row of Fig. \ref{fig: motivation}. (a). A possible reason is that when sticking to the trajectory context, a route will be completed by using the road segment that frequently co-occurs and thus skipping the inputs that less commonly co-occur due to being far or involving turns.
High-complexity trajectories could account for 30\% of the dataset. Thus, it deserves more attention.

To this end, this work {aims for a more robust trajectory recovery when dealing with complex trajectories. Moreover, the trajectory complexity could also affect other trajectory-related tasks such as route planning, next location prediction, and travel time estimation. To encourage model generalization, a pre-trained model is desired to learn a more robust road segment representation for complex trajectories in  different trajectory-related tasks}. 

Firstly, we define the \textit{complexity} of a trajectory from two aspects: (1) the Detour Score (\textit{DS}) to measure how far a trajectory derails from the shortest path between origin and destination \cite{barthelemy2011spatial} and (2) the Entropy Score (\textit{ES}) to measure the uncertainty from the turns that a trajectory makes \cite{xie2007measuring}, with details in Def. 3.

To overcome the limitations of trajectory contexts, we propose to consider not only the contexts of the trajectory, but also the \textit{semantics} of the road network. The road network semantics formulate the pairwise relations between two nodes in the road networks, usually in a graph form. {Intuitively, the neighbors on the road network should have similar representations}. Unlike the traditional graphs that are defined based on road network's static properties such as Euclidean distance, connectivity \cite{geng2019spatiotemporal}, and Point of Interest (POI) \cite{li2020tensor}, we define a multi-view semantic graph that is aware of the trajectory complexity based on the dynamic mobility. Then, we propose two modules to ensure the model robustness over high trajectory complexity: (1) in the trajectory aspect, we design a \textit{trajectory-dependent} multi-view graph aggregator, to adaptively aggregate the multi-view graph semantics according to trajectory contexts; (2) in the road segment aspect, we design a \textit{complexity-aware} Transformer that pays higher attention to the critical nodes when dealing with high-complexity trajectory. {These two designs allow our model to be critical perceptual.} To this end, the proposed model is named Multi-view Graph and Complexity Aware Transformer (\textit{MGCAT}).

To the best of our knowledge, it is the first work dedicated to tackling the complex trajectory pre-training. The main contributions of this paper are summarized as follows: 
\begin{itemize}
    \item  Define trajectory complexity and propose a Multi-view Graph and Complexity Aware Transformer model, which is more robust for complex trajectory pre-training.
    \item Define road network multi-view graph based on trajectory complexity and design a trajectory-dependent graph encoder, aggregated adaptively to different trajectories. Moreover, we design a complexity-aware transformer with extra attention to the nodes that are far or making turns in complex trajectories.
    \item Conduct prudent experiments to prove increasing better recovery when the trajectory gets more complex.
\end{itemize}

\section{Related Work}
\label{sec: lr}
We will introduce the related work about downstream trajectory recovery task and the fundamental trajectory modeling.

\subsection{Trajectory Recovery}
Existing trajectory recovery works try to tackle that trajectories are recorded at a low sampling rate. Traditional methods rely on probabilistic models such as Hidden Markov Model \cite{newson2009hidden} or heuristic search algorithms \cite{zheng2015trajectory, wu2016probabilistic}. For example, \citeauthor{wu2016probabilistic} proposed a pure probabilistic model with a temporal model via matrix factorization, a spatial model via inverse reinforcement learning, and a final route search to maximize posterior probability. Those methods have been proved flawed since they fail to capture complex sequential dependencies nor offer a high generality as pre-training models. 

Sequence-to-sequence methods have been proposed, \textit{e.g.}, DHTR \cite{w:19} and MTrajRec \cite{MTrajRec}. DHTR recovers a high-sampled trajectory from a low-sampled one in the free space, with a post-calibration Kalman Filter to reduce the predictive uncertainty. MTrajRec aligns the off-the-map input records back to the road, and then recovers an entire trajectory, so it targets the multi-task of map-matching and recovering from aligning with the road network, with map constraints in the output layer. Unlike MTrajRec solving a general problem where the inputs are not aligned with the map due to the GPS error, we are solving a more advanced problem where the complex trajectory couldn't be fully recovered on top of the input that is already being map-matched.

\subsection{Trajectory Modeling}
Sequence modeling has been successfully applied in natural language processing (NLP). 
Trajectory is also typical sequence data, and it shares many common characteristics with sentences in NLP. Therefore, most research is devoted to modelling the trajectory and its context by borrowing natural language models. For example, DeepMove \cite{z:18} implemented Word2vec \cite{2013Efficient} to model human movements between locations. Traj-RNN \cite{wu:17} leveraged the strength of recurrent neural network (RNN) to model the road segment sequence to consider long-term context information. TULVAE \cite{zhou2018trajectory} and TrajVAE \cite{chen2021trajvae} proposed neural generative architecture, \textit{i.e.}, Variational AutoEncoder (VAE), with stochastic latent variables that span hidden states in RNN and LSTM, respectively. 
However, all of these methods only model the trajectory itself or incorporate the topological constraints into the softmax layer in a hard manner. It is ignored to capture the influence of graph semantics on trajectory sequences and versa vise. As a result, the methods could not work well on complex trajectories.

To better incorporate graph semantics, 
Trembr \cite{fl:20} proposed to learn trajectory embeddings with road networks in a two-stage learning manner: it first uses Road2Vec to learn road segment representations based on road network semantics, then Traj2Vec to learn trajectory representation based on trajectory contexts. This may suffer from error propagation.
The most recent and related work is CTLE \cite{l:21}, which mainly aims at representation learning for multi-functional locations. It assumes that the same location can serve different functions at different times. Thus, the contribution of CTLE lies in consideration of trajectory context and time-aware features. However, it is not enough in complex trajectory situations. As we showed in Fig. \ref{fig: motivation}, it fails to handle complex trajectories since it ignores the informative road network graph semantics, and fails to be aware of the critical nodes in trajectories.


\section{Preliminaries and Problem Definitions}
\label{sec: prem}
We will give important definitions and problem statement. Throughout this exposition, scalars are denoted in italics, \textit{e.g.}, $n$; vectors by lowercase letters in boldface, \textit{e.g.}, $\mathbf{z}$; and matrices by uppercase boldface letters, \textit{e.g.}, $\mathbf{X}$; Sets by boldface script capital, \textit{e.g.}, $\mathcal{G}$.
\subsection{Preliminaries}
\textbf{Definition 1: \emph{Trajectory (Input)}.} A trajectory $\mathbf{S}$ is a sequence of spatiotemporal records with road segment $s_i$ and its timestamp $t_i$: $\mathbf{S} = \left \{ (s_1, t_1), ..., (s_i, t_i), ...( s_{\left | \mathbf{S} \right |}, t_{\left | \mathbf{S} \right |}) \right \}$. $\mathbf{S}$ is low-sampled and discrete. Besides, $\mathbf{S}$ usually needs to be matched to the map first \cite{lou2009map}.

\textbf{Definition 2: \emph{Recovered Trajectory (Output)}.} A recovered trajectory $\boldsymbol{\tau}$ is a complete and continuous trajectory that passes $\mathbf{S}$. $\boldsymbol{\tau}$ is a high-resolution version of $\mathbf{S}$. 

\textbf{Definition 3: \emph{Trajectory Complexity}.} As one of our technical contributions, trajectory complexity is defined with detour score (\textit{DS}) and entropy score (\textit{ES}):
\begin{equation}
    \begin{aligned}
         \textit{DS}(\mathbf{S}) &= \frac{RL(\mathbf{S})}{RL(\mathbf{S}^*_{[od]})}, \\
         \textit{ES}(\mathbf{S}) &= -\frac{1}{|\mathbf{S}|}\sum_{s_i \in \mathbf{S}} \sum_{{s'_i \in \mathcal{X}_i}}{P(s_i, s'_i) \log P(s_i, s'_i)}. \\
    \end{aligned}
\label{eq: dses}
\end{equation}

When computing \textit{DS}, $\mathbf{S}^*_{[od]}$ is the shortest path between origin and destination, and $RL(\cdot)$ is the route length of a trajectory. When computing \textit{ES},  $\mathcal{X}_i$ is the set of road segments that intersect with the segment $s_i$, $P(s_i, s'_i)$ is the flow transition probability from $s_i$ to $s'_i$, {calculated as the ratio of the flow of $s_i \to s'_i$ to the overall outflow from $s_i$}, and $|\mathbf{S}|$ is the number of road segments in $\mathbf{S}$. Generally, as shown in Fig. \ref{fig: motivation}. (b1), a higher \textit{DS} means a more derailing trajectory ($\textit{DS}_{\textit{traj-1}} > \textit{DS}_\textit{traj-2}$). As shown in Fig. \ref{fig: motivation}. (b2), a higher \textit{ES} means a more zig-zag trajectory with higher uncertainty ($\textit{ES}_{\textit{traj-3}} > \textit{ES}_\textit{traj-1}$, yet $\textit{DS}_{\textit{traj-3}} = \textit{DS}_\textit{traj-1}$). The final complexity of $\mathbf{S}$ is the weighted combination of the normalized \textit{DS} and \textit{ES}: 
\begin{equation}
\label{eq: complexity}
    \textit{complexity}_{\mathbf{S}} = \mu \cdot \textit{DS}(\mathbf{S})+\nu \cdot \textit{ES}(\mathbf{S})
\end{equation}
where  $\mu$ and $\nu$ are the weights.

\textbf{Definition 4: \emph{Trajectory Contexts}.} The trajectory context of $s_i \in \mathbf{S}$ is the spatiotemporal information and complexity of the whole trajectory $\mathbf{S}$, with our formulation in Eq. (\ref{eq: q_S}).

\textbf{Definition 5: \emph{Road Network Graph (Semantics)}.} The road network graph $\mathcal{G}(\mathcal{V}, \mathcal{E})$ is defined as a directed graph, where $\mathcal{V}$ is the set of $|\mathcal{V}|$ road segments, and $\mathcal{E}$ is the set of edges. Two graphs are introduced, \textit{i.e.}, a route distance graph $\mathcal{G}^{d}$ and a route entropy graph $\mathcal{G}^{e}$, with details in Eq. (\ref{eq: G_d}) and (\ref{eq: G_e}).

\subsection{Problem Statement}

Given a low-sampled trajectory $\mathbf{S}$ and road network multi-view graph $\mathcal{G}$, we aim to recover a complete trajectory $\boldsymbol{\tau}$ in high resolution: $\boldsymbol{\tau} \sim \textit{MGCAT}(\mathbf{S}, \mathcal{G})$ . It contains two steps:
\begin{itemize}
    \item Pre-training: to learn a mapping function $f(\cdot)$ that projects the road segment from $\mathbf{S}$ to a latent embedding vector $\mathbf{z} \in \mathbb{R}^{d_h}$, where $d_h$ is the embedding space dimension: $\mathbf{Z}_{S} \sim f(\mathbf{S})$, where $\mathbf{Z}_{S} \in \mathbb{R}^{\left | \mathbf{S} \right | \times d_h}$.
    \item Fine-tuning for downstream task: to recover the complete trajectory $\boldsymbol{\tau}$ based on $\mathbf{Z}_{S}$: $\boldsymbol{\tau} \sim \textit{Dec}(\mathbf{Z}_{S})$, where $\textit{Dec}(\cdot)$ is a decoder. 
\end{itemize}

\begin{figure}[t]
\centering
\includegraphics[width=\columnwidth]{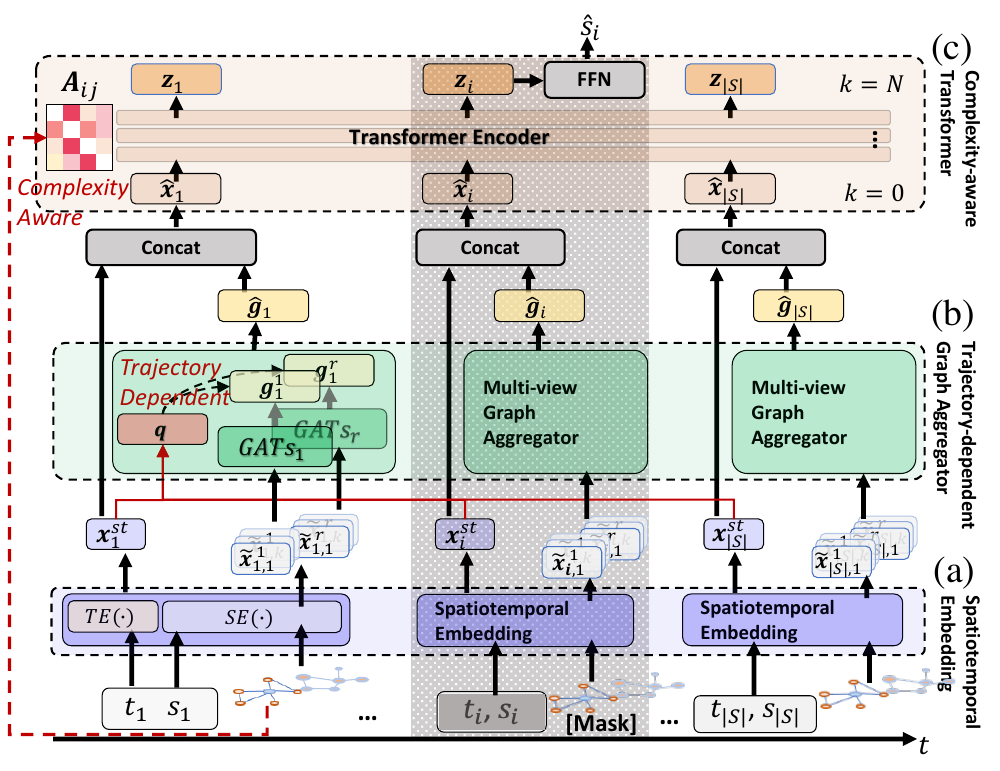} 
\caption{The proposed MGCAT, with modules: (1) Spatiotemporal Embedding Layer (in color blue), (2) Trajectory-dependent Multi-view Graph Aggregator (in color green), and (3) Complexity-aware Transformer (in color orange). Our two critical designs, \textit{i.e.}, trajectory-dependent factor $\mathbf{q}_{\mathbf{S}}$ and complexity-aware attention $\mathbf{A}$ are highlighted in red.}
\label{fig_framework}
\end{figure}

\section{Model Framework}
\label{sec: model}
The trajectory pre-training framework MGCAT consists of three modules, as shown in Fig. \ref{fig_framework}: (1) A shared spatiotemporal embedding layer; (2) A trajectory-dependent multi-view graph aggregator, with an adaptive weight for each individual trajectory and its complexity; (3) A pre-training module based on a complexity-aware Transformer to pay extra attention to the critical nodes in complex trajectories.

\subsection{Spatiotemporal Embedding Layer}
\label{sec: ST_embed}
Unlike the sequence in the language model, the trajectory $\mathbf{S}$ is with typical spatiotemporal attributes. Both the information ``where'' and ``when'' are important. 
Specifically, given an input road segment sequence $\mathbf{S} = \left \{ (s_1, t_1), ..., ( s_{\left | \mathbf{S} \right |}, t_{\left | \mathbf{S} \right |}) \right \}$, we obtain spatial embedding and temporal encoding as follows:
\begin{align}\small
    \mathbf{x}_i^s = \textit{SE}(s_i), \  \mathbf{x}_i^t = \textit{TE}(t_i),
\end{align}
where $\textit{SE}(\cdot)$ is a spatial token embedding layer such as word embedding (\textit{e.g.}, \textit{word2vec}) in BERT \cite{m:13}. $\textit{TE}(\cdot)$ is a temporal encoding layer proposed by TGAT \cite{TGAT} that specifically embeds the unevenly-recorded timestamps on the temporal axis:
\begin{equation}\small
\small
\textit{TE}(t) = \\ \frac{1}{\sqrt{d_e}} [\cos(w_1t), \sin(w_1t), \dots, \cos(w_{\frac{d_e}{2}}t), \sin(w_{\frac{d_e}{2}}t)].
\end{equation}

The $\textit{TE}(\cdot)$ is similar to the positional encoding in the canonical Transformer \cite{v:17}, except the absolute visited timestamp $t$ is used instead of the position index. $\big\{ w_1, ..., w_{\frac{d_e}{2}} \big\}$ are trainable parameters, and $d_e$ is the latent space dimension in embedding layers. 

The spatial and temporal embeddings for trajectory $\mathbf{S}$ are denoted in matrix forms, \textit{i.e.}, $\mathbf{X}^s, \mathbf{X}^t \in \mathbb{R}^{\left | \mathbf{S} \right | \times d_e}$, respectively. The final spatiotemporal embedding is defined as $\mathbf{X}^{st} = \mathbf{X}^s + \mathbf{X}^t$, where $\mathbf{X}^{st} \in \mathbb{R}^{\left | \mathbf{S} \right | \times d_e}$.

\subsection{Trajectory-dependent Multi-view Graph Aggregator}

However, it is insufficient to only utilize spatiotemporal information for segment embedding. As we highlighted before, the road network offers abundant graph semantics for segment embedding. The semantic graphs of road networks play a significant role in human mobility, \textit{e.g.},  the topological graph and Point of Interest (POI) graph\cite{geng2019spatiotemporal,li2020tensor,ziyue2021tensor}, where neighbors are assumed to have a similar pattern. Unlike the traditional graph definition, we aim to construct graphs that can reflect the trajectory complexity in fidelity. 1) We first construct multi-view graph based on trajectory complexity, and 2) then conduct graph embedding, and 3) finally aggregate the multi-views in an adaptive and trajectory-dependent manner.

\subsubsection{Multi-view Graph Construction}
\label{sec: multi-view-graph}
Two graphs reflecting the complexity will be constructed.

\textbf{Route distance graph}: Unlike the simple shortest Euclidean distance between two road segments, which is in an over-ideal and unpractical setting, we utilize the actual route distance traveling from segment $s_i$ to $s_j$ as it contains the authentic information of how difficult to commute between the two, thus considering the detour factor in the complexity definition.
We construct the route distance graph $\mathcal{G}^{d}$ and denote the route distance adjacency matrix as $\mathbf{J}^d \in \mathbb{R}^{|\mathcal{V}| \times |\mathcal{V}|}$:
\begin{equation}\small
\label{eq: G_d}
    \mathbf{J}^d_{ij} = RL(\bar{\mathbf{S}}_{k,[ij]}).
\end{equation}

Since there are multiple routes between two segments, $\bar{\mathbf{S}}_{k,[ij]}$ is designed as the average of $k$ randomly selected routes from $s_i$ to $s_j$. 
$\mathbf{J}^d_{ij}$ measures how hard to commute between $s_i$ and $s_j$ in reality. 

\textbf{Route entropy graph}: Similar to the entropy score in Eq. (\ref{eq: dses}) to measure the uncertainty of a trajectory when making turns, 
a route choice entropy graph $\mathcal{G}^{e}$ is also designed this way for two road segments. The entropy adjacency matrix $\mathbf{J}^e \in \mathbb{R}^{|\mathcal{V}| \times |\mathcal{V}|}$ is defined to measure the uncertainty accumulated along traveling from $s_i$ to $s_j$ along the trajectory $\mathbf{S}_{[ij]}$:

\begin{equation}\small
    \begin{aligned}
        \mathbf{J}^e_{ij} =  -\sum_{s_i \in \mathbf{S}_{[ij]}} \sum_{{s'_i \in \mathcal{X}_i}}{P(s_i, s'_i) \log P(s_i, s'_i)}. \\
    \end{aligned}
\label{eq: G_e}
\end{equation}

For both of the graph adjacency matrices, in practice, we fill in a large value if no route exists and the matrix is normalized along the column axis with the min-max scaler. 

\subsubsection{Graph Embedding}
\label{sec: GAT}

As mentioned before, the multi-view graph provides semantics, namely, the pairwise relation of two road segments. This can be encoded via graph embedding. Instead of embedding a segment $s_i$ solely, the information of its neighboring nodes $\mathcal{N}_i$ can help more robust trajectory representation learning, because neighbors could be the alternatives of $s_i$ for a trajectory. Introducing such neighborhood information could help more robust representation as explored in \textit{NNCLR} \cite{dwibedi2021little}. Thus, we apply 
Graph Attention Networks (GATs)  \cite{velivckovic2017graph} to encode the road segment $s_i$ in the $r$-th graph $\mathcal{G}^r$ with its neighbors $\mathcal{N}_i^r$. 
As shown in Fig. \ref{fig_framework}, road segment $s_i$ along with its first-order neighbors in $r$-th graph go through the same $\textit{SE}(\cdot)$, so that we obtain  $\{\tilde{\mathbf{x}}_{i,i'}^r\}_{i'\in \mathcal{N}_i^r} ^{r \in R}$ as its neighbor semantics. Then, they are input together into \textit{GAT}s:
\begin{align}\small
    \mathbf{g}^r_i = \textit{GATs}(\mathbf{x}_i^s, \mathcal{G}^r),
\end{align}
where $\mathbf{g}^r_i \in \mathbb{R}^{d_e}$, $r \in R$, and $R$ is the total amount of views (in our case, $R=2$, they are $\mathbf{J}^d$ and $\mathbf{J}^e$ defined above).
\textit{GAT}s is effective and spatially robust, and easy to be co-trained with Transformer \cite{velivckovic2017graph}.

\subsubsection{Trajectory-dependent Multi-view Graph Aggregator}
\label{sec: aggregator}
When conducting graph embedding for road segments, \textit{GAT}s only utilizes the information of road network's graphs, \textit{i.e.}, graph semantics. It ignores that the road segments are also the elements of trajectory sequences, namely, trajectory context is ignored.
However, it is essential to incorporate trajectory contexts into road network representation \cite{chen2021robust}. 
Similar to the natural language where the same word could mean differently in different sentences, the same graph node should get a different ultimate graph embedding when it appears in a different trajectory. Thus, we propose to introduce trajectory context when aggregating the multi-view embeddings of road segment $s_i$. By this design, the graph aggregator is achieved as \textit{trajectory-dependent}. 

We define the trajectory context $\mathbf{q}_{\mathbf{S}} \in \mathbb{R}^{d_e}$ 
as the average spatiotemporal feature of the trajectory $\mathbf{S}$ scaled by its complexity:
\begin{align}\small
\label{eq: q_S}
    \mathbf{q}_{\mathbf{S}} = \textit{complexity}_{\mathbf{S}} \cdot \frac{1}{|\mathbf{S}|} \sum_{i \in |\mathbf{S}|}\mathbf{x}_i^{st}.
\end{align}

{As a result, this $\mathbf{q}_{\mathbf{S}}$ contains all the information we value, \textit{i.e.}, the complexity and the spatiotemporal property.}

Then, the trajectory context is introduced as a bias term into the attention coefficient of the $i$-node in the $r$-th view, denoted as $w_{i, \mathbf{S}}^r$. {In such design, the node from a more spatiotemporal complex trajectory might gain higher attention}:

\begin{align}\small
\label{w}
    w_{i, \mathbf{S}}^r = \mathbf{v}^\top \sigma (\mathbf{W}^r \mathbf{g}_i^{r} + \mathbf{W} \mathbf{q}_{\mathbf{S}}),
\end{align}

\noindent where $\sigma(\cdot)$ is an activation function, \textit{e.g.}, $\textit{tanh}$. $\mathbf{v} \in \mathbb{R}^{d_e}, \mathbf{W}^r$, $\mathbf{W} \in \mathbb{R}^{d_e\times d_e}$ are the trainable weight vectors, the weight matrix for $r$-th view, and a shared weight matrix, respectively. 

Finally, the attention score $\alpha_{i, \mathbf{S}}^r$ on graph embedding $\mathbf{g}_i^r$ is designed as a soft-max function:

\begin{align}\small
\label{alpha}
    \alpha_{i,\mathbf{S}}^r = \frac{\exp(w_{i, \mathbf{S}}^r)}{\sum_{r' \in R} \exp(w_{i, \mathbf{S}}^{r^{'}})}.
\end{align}

The graph embeddings are aggregated as a linear weighted sum of all the views in a trajectory-dependent way:

\begin{align}\small
    \hat{\mathbf{g}}_{i,\mathbf{S}} = \sum_{r \in R} \alpha_{i,\mathbf{S}}^r \cdot \mathbf{g}_i^{r}. 
\end{align}

As a result, trajectories have different weights on different views of the graph, due to the trajectory contexts considered. 

\subsection{Complexity-Aware Pre-training}
\label{sec: transformer}
The trajectory pre-training module is designed to fully consider the spatiotemporal property and graph semantics. 
Thus, we first hybridize the input by concatenating the spatiotemporal embedding $\mathbf{x}_i^{st}$ and multi-view graph embedding $\hat{\mathbf{g}}_{i,\mathbf{S}}$. It is worth mentioning that the $\mathbf{x}_i^{st}$ is the general spatiotemporal property that is trajectory-independent as long as $s_i$ and $t_i$ are given, yet $\hat{\mathbf{g}}_{i,\mathbf{S}}$ is trajectory-dependent, considering trajectory contexts and graph semantics.

The hybrid embedding $\hat{\mathbf{x}}_i \in \mathbb{R}^{d_h}$ is formulated as:
\begin{align}\small
    \hat{\mathbf{x}}_i = [\mathbf{W}^{st}\mathbf{x}_i^{st} || \mathbf{W}^{g}\hat{\mathbf{g}}_{i, \mathbf{S}}],
\end{align}
where $\mathbf{W}^{st}, \mathbf{W}^{g} \in \mathbb{R}^{ \frac{{d_h}}{{2}} \times d_e}$ are learnable parameters and $||$ represents concatenation.

The hybrid embedding $\hat{\mathbf{x}}_i$ is then input into Transformer encoder layers, and for the $k$-th layer:
\begin{equation}\small
\begin{aligned}
    \big \{ \mathbf{h}_1^{(k)}, ..., \mathbf{h}_{\left | \mathbf{S} \right |}^{(k)} \big \} =  \textit{TransEnc}(\big \{ \mathbf{h}_1^{(k-1)}, ..., \mathbf{h}_{\left | \mathbf{S} \right |}^{(k-1)} \big \}),
\end{aligned}
\end{equation}
where $\textit{TransEnc}$ represents the Transformer encoder layer. The $0$-th layer is $\big \{ \mathbf{h}_1^{(0)}, ..., \mathbf{h}_{\left | \mathbf{S} \right |}^{(0)} \big \} = \left \{ \hat{\mathbf{x}}_1, ..., \hat{\mathbf{x}}_{\left | \mathbf{S} \right |} \right \}$, and the $N$-th layer is the output $\big \{ \mathbf{h}_1^{(N)}, ..., \mathbf{h}_{\left | \mathbf{S} \right |}^{(N)} \big \} = \big \{ \mathbf{z}_1, ..., \mathbf{z}_{\left | \mathbf{S} \right |} \big \}$.  

\textbf{Objective Function:} During pre-training, we mask some road segments and adopt the Masked Language Model (MLM) objective function in BERT to achieve generality \cite{d:18}.

\textbf{Complexity-aware Attention via Soft-mask Term:} However, as demonstrated in Fig. \ref{fig: motivation}, the traditional transformer-based methods such as CLTE \cite{l:21} mistakenly skip several critical road segments when the trajectory is complex. We believe the reason is that when dealing with complex trajectories, the transformer fails to pay enough attention to those segments that are too far away or involve turns, as they are less likely to co-occur with the rest of the training corpus. The claim will be further validated in our experiment, with results shown in Fig. \ref{fig_attention_reasons}.

Motivated by Graphormer \cite{Graphormer}, which introduced a soft-mask term $b_{\phi(s_i, s_j)}$ to the attention to incorporate graph structural information, where $\phi(s_i, s_j)$ measures the spatial relation between $s_i$ and $s_j$. 
We design a different soft-mask so as: (1) to pay \textbf{more} attention to the critical nodes with \textbf{higher} route distance or entropy value \textbf{only} when the road trajectory is complex. This is opposite to the common attention that is paid less to the nodes that are far away; (2) to stay inactive when dealing with average (i.e., low and middle complexity) trajectories in case the model is misled by the opposite attention. {To achieve the different attention direction when dealing with highly-complex trajectories only, we will introduce a bias term based on an indicator function to the attention soft-mask term.} 

Thus, as shown in Eq. (\ref{eq: redesigned_attention}), two $b_{\phi(s_i, s_j)}$s are introduced: 
$\phi(s_i, s_j)$ are chosen as the two adjacency matrices $\mathbf{J}^d$ and $\mathbf{J}^e$. Besides, our $b_{(\cdot)}$ is a learnable \textbf{increasing} function, achieving more attention to the nodes that are far away. Moreover, we design the soft-mask term as an indicator function, which returns $b_{\mathbf{J}_{ij}^d} + b_{\mathbf{J}_{ij}^e}$  only if the complexity is higher than a threshold $\theta$; Otherwise, the term is zero. As a result, the attention achieves being \textit{complexity-aware}.

\begin{equation}\small
\label{eq: redesigned_attention}
    \begin{aligned}
        & \mathbf{A}_{ij} = \frac{(\mathbf{h}_i\mathbf{W}^Q)(\mathbf{h}_j\mathbf{W}^K)^\top}{\sqrt{d_k} } + \mathbf{1}_{ \{\text{\textit{complexity}}>\theta \} } \left(b_{\mathbf{J}_{ij}^d} + b_{\mathbf{J}_{ij}^e}\right),\\
    \end{aligned}
\end{equation}
where $\mathbf{W}^Q, \mathbf{W}^K \in \mathbb{R}^{d_h \times d_k}$ are trainable weights,  $d_k = d_h / m$, $m$ is the number of multi-head, $\mathbf{1}(\cdot)$ is an indicator function. 

\section{Experiment and analysis}
\label{sec: experiment}
In this section, we conduct extensive experiments on three real-world trajectory datasets and provide detailed analyses to demonstrate the improvements of the proposed MGCAT. 

\subsection{Datasets}
We use three real-world taxi GPS trajectory open datasets from the city Xi'an (XA), Chengdu (CD), and Porto (PT). Xi'an and Chengdu datasets are released by Didi\footnote{https://outreach.didichuxing.com/}
from 01-Oct-2016 to 30-Nov-2016. The Porto dataset\footnote{https://www.kaggle.com/c/pkdd-15-predict-taxi-service-trajectory-i/data}
is provided by Kaggle with 442 taxis from 07-Jan-2013 to 30-Jun-2014. We extract trajectories by order ID and map-match raw GPS trajectories to the road network with 
a hidden Markov model-based algorithm \cite{newson2009hidden}. The datasets are summarized in Table~\ref{tab: data_summary}.

\begin{table}[ht]
\caption{The summary of the datasets.}
\centering
\begin{tabular}{c|ccc}
\toprule
& Xi'an   & Chengdu & Porto   \\ \hline
\# of road segments $|\mathcal{V}|$ & 6,161   & 6,566   & 11,237  \\
\# of crossroads    & 2,683   & 2,848   & 5,258   \\
\# of trajectories  & 200,000 & 80,000  & 180,000 \\
Avg. length $\bar{|\mathbf{S}|}$ & 24.35 & 12.30 & 41.54 \\ 
Avg. $\textit{complexity}$ & 0.390 & 0.448 & 0.464 \\
\bottomrule
\end{tabular}
\label{tab: data_summary}
\end{table}

\subsection{Benchmark Methods}
We compare our method with the following benchmarks. 

\begin{itemize}
\item \textbf{Road2Vec} \cite{fl:20}. Road2Vec embeds road segments into vectors by treating the trajectory as a sequence, and uses a sliding window to capture context pairs along a trajectory.

\item \textbf{Traj-RNN} \cite{wu:17}. Traj-RNN, also known as Traj2vec, employs the RNN \cite{h:97} to model the temporal road segment sequence and incorporates spatial topology as constraints via a state-constrained softmax function. 

\item \textbf{Trembr} \cite{fl:20}. Trembr combines Road2Vec to learn road segment embedding and Traj2vec to learn trajectory embedding.

\item \textbf{Traj-VAE} \cite{chen2021trajvae}. 
Traj-VAE uses LSTM to model trajectories and employs the VAE to learn the distributions of latent random variables of trajectories. 

\item \textbf{CTLE} \cite{l:21}. CTLE implemented BERT to model trajectories to consider temporal features along with trajectory context. 
\end{itemize}

\subsection{Evaluation Metrics}

We evaluate trajectory recovery quality from two aspects: 

(1) \textbf{Classification metrics} to compare two sets of road segments, \textit{i.e.}, Precision $P$, Recall $R$, and F1-score $F1$, all of which are the higher, the better.

(2) \textbf{Geographic distance metrics} to compare two continuous trajectories \cite{s:20}, \textit{i.e.}, One Way Distance (OWD) and Merge Distance (MD). OWD measures the average integral of the following two distances: 1) the distance from points of trajectory-1 to trajectory-2 divided by the length of trajectory-1, and 2) oppositely the distance from trajectory-2 to trajectory-1. MD measures the shortest super-trajectory 
that connects all the sample points from the recovered trajectory-1 and the input trajectory-2. The smaller distance metrics indicate better performance. 

\subsection{Experiment Settings}
\textbf{Data Preparation:} (1) For each dataset, we set 80\% and 20\% as the pre-training set and downstream task set, respectively. 
(2) For the pre-training, we randomly mask 2/3 of the original road segments from a sequence with [MASK] token, and leverage MLM objectives to learn the parameters.
(3) For the trajectory recovery, we randomly select 2/3 of the original road segments from each sequence as inputs and use the embedding generated by the pre-trained model to recover the complete trajectory. 
(4) We use all the trajectories in the pre-training set to construct the multi-view graph.
(5)  Threshold $\theta$ in Eq. (\ref{eq: redesigned_attention}) can be set as the 75\% quartile of the distribution.  Complexity weights $\mu$ and $\nu$ in Eq. (\ref{eq: complexity}) are 0.5 if without prior knowledge. 

\textbf{Pre-training Model:} For our MGCAT, we stack 2 Transformer encoder layers with 8 attention heads. The hidden feature size is set to 512, and the size of all embedding vectors is set to 128. We use the graph attention network with 2 attention heads, and the hidden feature size is set to 256. We train the pre-trained model for 10 epochs with the batch size as 16. We choose Adam optimizer \cite{kingma2014adam} with an initial learning rate as $10^{-4}$.

\textbf{Downstream Model:} For the downstream model, we simply implement the sequence model based on the Gated Recurrent Unit as a decoder with 2 recurrent layers and 256 features in the hidden state, which is the standard model for sequence modeling and can better reflect the embedding of the pre-trained model quality. The decoder terminates the generated sequence according to the EOS tag or a limited length. We use the early stopping mechanism \cite{yao2007early} to train the downstream task model with the loss function set as the cross-entropy loss. 

The neural networks are all implemented using Pytorch and optimized by the Adam method. All experiments are conducted on a workstation with an Intel(R) Core(TM) i7-8700 CPU @ 3.20GHz, and NVIDIA RTX1080ti GPUs.

\begin{table}[t]
\centering
\caption{Trajectory recovery performance comparison of different approaches on three real-world datasets.}
\small
\begin{tabular}{c|c|ccc|cc}
\toprule
City & Method & \begin{tabular}[c]{@{}c@{}}P{\tiny (\%)}\end{tabular} & \begin{tabular}[c]{@{}c@{}}R{\tiny (\%)}\end{tabular} & \begin{tabular}[c]{@{}c@{}}F1{\tiny (\%)}\end{tabular} & OWD & MD  \\ \hline
\multirow{5}{*}{PT} & Road2Vec & 52.92 & 52.53 & 52.72 & 68 & 0.2390  \\
 & Traj-RNN & 54.56 & 55.20 & 54.88 & 109 & 0.2564 \\
 & Trembr & 55.96 & 55.43 & 55.69 & 110 & 0.2533 \\
 & Traj-VAE & 54.34 & 55.88 & 55.10 & 132 & 0.2986 \\
 & CTLE & \underline{57.07} & \underline{59.53} & \underline{58.27} & \underline{62} & \underline{0.2164} \\
 & \textbf{Our} & \textbf{58.82} & \textbf{61.24} & \textbf{60.01} & \textbf{47} & \textbf{0.1809}  \\ \hline
\multirow{5}{*}{CD} & Road2Vec & 53.60 & 52.74 & 53.03 & 274 & 0.3786 \\
 & Traj-RNN & 57.59 & 56.75 & 57.17 & 269 & 0.3376 \\
 & Trembr & 57.86 & 57.15 & 57.51 & 273 & 0.3391 \\
 & Traj-VAE & 58.74 & 58.63 & 58.68 & 257 & 0.3496 \\
 & CTLE & \underline{59.81} & \underline{60.66} & \underline{60.23} & \underline{67} & \underline{0.1763}\\
 & \textbf{Our} & \textbf{63.41} & \textbf{63.09} & \textbf{63.25} & \textbf{56} & \textbf{0.1661} \\ \hline
\multirow{5}{*}{XA} & Road2Vec & 64.82 & 65.00 & 64.91 & 77 & 0.1982 \\
 & Traj-RNN & 66.44 & 66.95 & 66.68 & 124 & 0.2286 \\
 & Trembr & 66.79 & 66.75 & 66.52 & 110 & 0.1972 \\
 & Traj-VAE & 67.78 & 68.38 & 68.03 & 119 & 0.2176 \\
 & CTLE & \underline{69.09} & \underline{69.71} & \underline{69.40} & \underline{51} & \underline{0.1414} \\
 & \textbf{Our} & \textbf{72.80} & \textbf{73.42} & \textbf{73.02} & \textbf{40} & \textbf{0.1166} \\ 
\bottomrule
\end{tabular}
\label{tab: all_comparison}
\end{table}

\subsection{Improved Performances of Our Model}
Table \ref{tab: all_comparison} compares the overall performances of different models, with the best result in boldface, second-best underlined. As the results demonstrate: (1) Road2Vec has the worst performance, given it only captures the co-occurrence information of road segments within the window size, ignoring the long-range dependency; (2) Traj-RNN and Trembr have  better performance than Road2Vec due to well-preserved  long-range dependency; Trembr has marginal gain over Traj-RNN as it combines Road2Vec and Traj2vec together; (3) Traj-VAE further learns the distributions of latent variables, with a more robust performance compared to Traj-RNN; (4) So far,  CTLE is the second-best method since it uses a trajectory to generate spatial and temporal embedding for the road segment, and feeds the embeddings into the attention, which helps to capture trajectory context better. However, CTLE's attention only reflects the spatiotemporal context from the trajectory, ignoring the multi-view road semantic graphs, and cannot handle the complex trajectories, which will be demonstrated in the following section. Our method achieves the best performance, with around 5\% higher F1 score and 17\% lower MD than~CTLE.

\subsection{When Dealing with Complex Trajectories}
We further scrutinize how the model performance varies to different trajectory complexities. We compare our method with the second-best method CTLE in the Xi'an dataset, with different trajectory complexity.

\begin{table}[t]
\centering
\caption{Performance comparison of our model and CTLE on three complexity levels of Low, Middle, and High, with the last row showing our \textbf{relative} gain compared to CTLE.}
\small
\begin{tabular}{c|ccc|ccc}
\toprule
Metric & \multicolumn{3}{c|}{F1{\tiny (\%)}} & \multicolumn{3}{c}{MD} \\ 
\hline
Level & Low & Mid & High & Low & Mid & High \\ 
\hline
CTLE & \underline{84.01} & \underline{70.82} & \underline{50.35} & \underline{0.0444} & \underline{0.1012} & \underline{0.2239} \\
\textbf{Our} & \textbf{86.29} & \textbf{73.41} & \textbf{54.46} & \textbf{0.0390} & \textbf{0.0871} & \textbf{0.1770} \\
& \scriptsize{\textit{+2.7\%}} & \scriptsize{\textit{+3.6\%}} & \scriptsize{\textit{+8.2\%}} & \scriptsize{\textit{-12\%}} & \scriptsize{\textit{-13.9\%}} & \scriptsize{\textit{-20.9\%}}
\\ \bottomrule
\end{tabular}
\label{tab: three_level_complexity_comparison}
\end{table}

\begin{figure}[t]
\centering
\includegraphics[width=0.99\columnwidth]{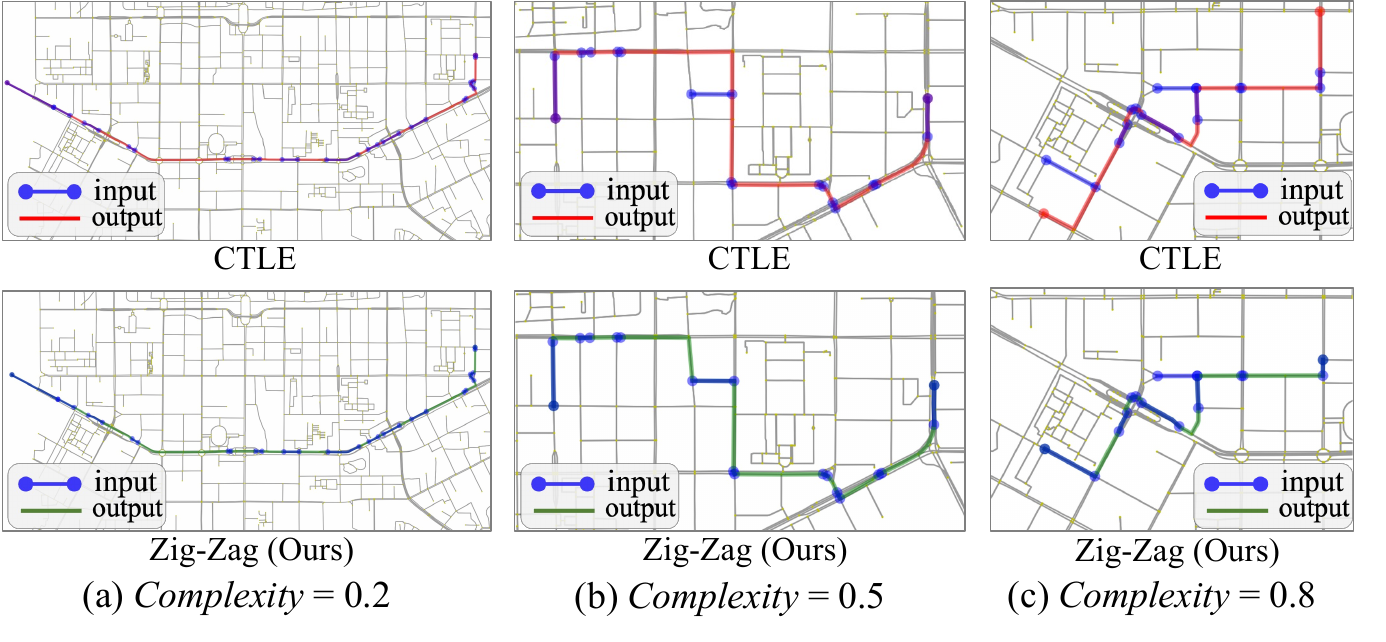} 
\caption{Trajectory Recovery Results of our MGCAT and CTLE on trajectories with different complexity}
\label{fig_three_recovered_trajectory}
\end{figure}

We set complexity lower than the 25\% quartile as ``Low'' complexity, 25\% to 75\% as ``Middle'' and higher than the 75\% quartile as ``High''. 
And we average the performance of our model and CTLE within three levels of Low, Middle, and High in Table \ref{tab: three_level_complexity_comparison}: As trajectory goes more complex, our model outperforms CTLE with increasing margin gain, \textit{i.e.}, 3.6\% higher F1, 13.9\% lower MD in Middle complexity, and 8.2\% higher F1, 20.9\% lower MD in High complexity. {We observe more performance gain for High complexity than Low and Middle complexity, this may be because our complexity-aware attention is designed with an indicator function, which will only be turned on for complex enough trajectories.} 

The intuitive examples for recovery results under different trajectory complexity are further visualized in Fig. \ref{fig_three_recovered_trajectory}, which provides a glimpse of the CTLE's and our model's performances over $\textit{complexity} = 0.2, 0.5, 0.8$ respectively. {As we could observe: (1) firstly, the complexity is defined quite reasonable, from $\textit{complexity} = 0.2$ to $0.8$, the trajectories are indeed more visually complex; (2) when $\textit{complexity} = 0.2$ (Low), CLTE could achieve the same correct result as MGCAT; however, when $\textit{complexity} = 0.5, 0.8$, CTLE starts to miss some inputs; but MGCAT could maintain the fidelity of the recovered trajectories.}

\subsection{How Our Attention Helps?}

\begin{figure}[t]
\centering
\includegraphics[width=0.99\columnwidth]{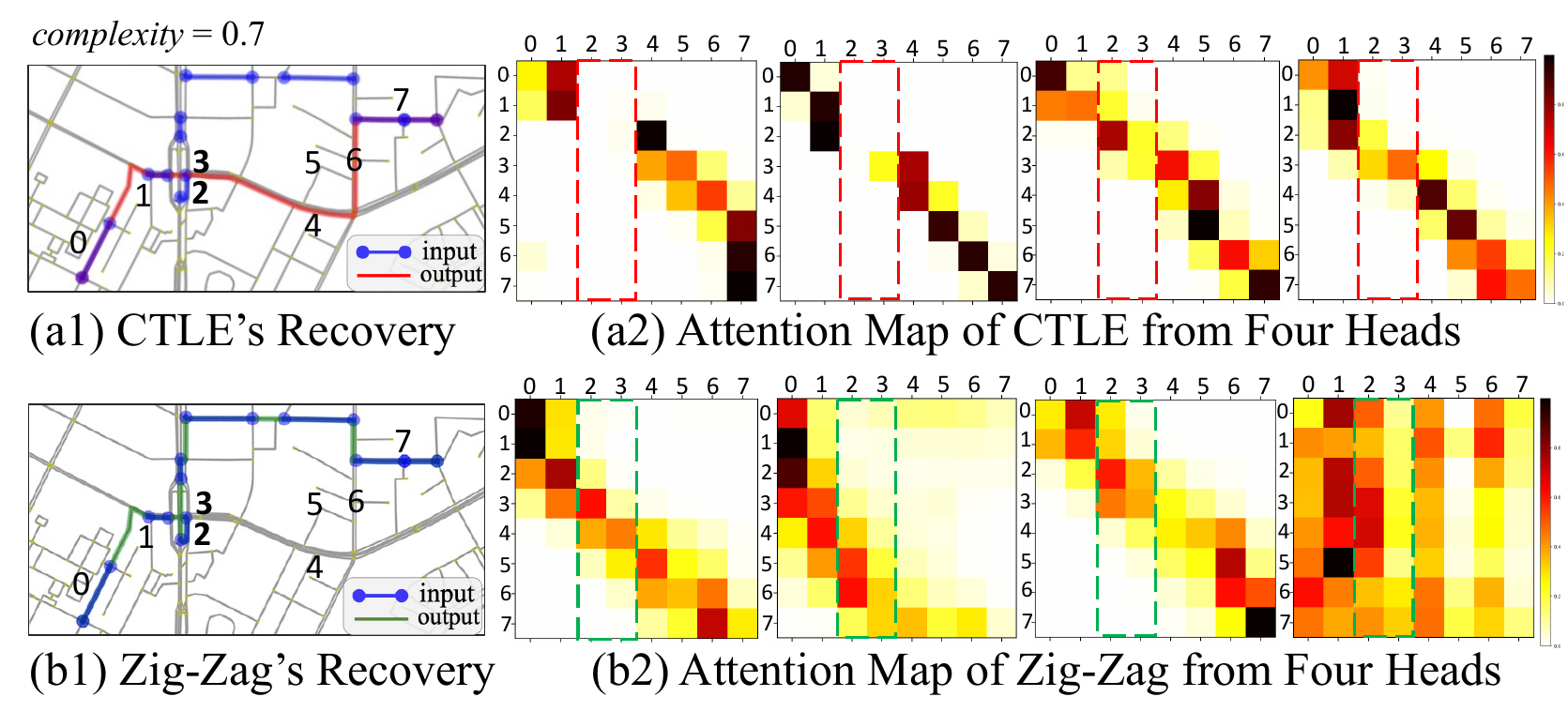} 
\caption{Attention maps for trajectory with \textit{complexity} = 0.7 from: (a) CTLE and (b) the proposed MGCAT}
\label{fig_attention_reasons}
\end{figure}

Here we try to answer why our MGCAT works better than the general transformer methods such as CTLE in complex trajectory recovery. We plot the four heads attention maps from  our complexity-aware Transformer and CTLE in Fig. \ref{fig_attention_reasons}, respectively. We observe two limitations of CTLE's attention: (1) Its attention mostly stays around the diagonal, meaning a node only focuses on itself or its close neighbors. This inevitably restricts the model to assume a simple and low-rank trajectory inherently; (2) Most importantly, when the trajectory goes complex, it loses attention on some road segments that are far or require turns.

As demonstrated in Fig. \ref{fig_attention_reasons}. (a1) and (b1), with $\text{complexity} = 0.7$, CTLE loses attention to segment ID-2 and 3 since segment ID-2 and 3 are making a U-turn. As a result, CTLE's recovered trajectory skips node ID-2 and 3 and chooses a popular main road for the recovery. Our enhanced attention preserves the multi-view semantics and goes wider than on the diagonal; The soft-mask is active when dealing with a complex trajectory and pays higher attention to node ID-2 and 3, as shown in Fig. \ref{fig_attention_reasons}. (b2). 

\section{Conclusion}
\label{sec: conclude}
This paper examines the problem of complex trajectory recovery. We define trajectory complexity and construct complexity-related multi-view graphs. Then we proposed a pre-training model, \textit{i.e.}, MGCAT: it has a trajectory-dependent multi-view graph aggregator that considers trajectory contexts and graph semantics, achieving robust embedding; it also has a complexity-aware Transformer with a well-designed soft-mask term to pay higher attention to critical nodes in complex trajectories. Such that, our MGCAT is critical perceptual to complex trajectories.

In future work, we plan to improve the semantic-aware ability of the model by incorporating geographic information (\textit{e.g.}, POI), and explore the extensive downstream applications of the trajectory pre-trained model, such as trajectory generation and trajectory-user linking tasks.

\bibliography{main}
\bibliographystyle{ACM-Reference-Format}

\vfill


\end{document}